\documentclass[journal,twoside,web]{ieeecolor}
\usepackage{generic}
\usepackage{cite}
\usepackage{amsmath,amssymb,amsfonts}
\usepackage{algorithmic}
\usepackage{graphicx}
\usepackage{textcomp}
\usepackage{url}

\usepackage{subfigure}
\usepackage{epstopdf}
\usepackage{color, soul}
\usepackage[dvipsnames]{xcolor}

\graphicspath{{./figures/}}

\newcommand{\N}{\mathbb{N}}
\newcommand{\mcl}[1]{\mathcal{#1}}

\newcommand{\R}{\mathbb{R}}

\def\BibTeX{{\rm B\kern-.05em{\sc i\kern-.025em b}\kern-.08em
    T\kern-.1667em\lower.7ex\hbox{E}\kern-.125emX}}

\begin{document}
\title{Employing Feature Selection Algorithms to Determine the Immune State of a Mouse Model of Rheumatoid Arthritis}

\author{ $^{1}$Aleksandr Talitckii, $^{1}$Joslyn L. Mangal,  $^{1}$Brendon K. Colbert, \\
$^{1}$Abhinav P. Acharya, $^{1}$Matthew M. Peet
\thanks{Manuscript received Month xx, 2023; revised Month xx, xxxx.}
\thanks{$^{1}$ School for the Engineering of Matter, Transport and Energy, Arizona State University, Tempe, AZ, 85298 USA  \\
E-mail: atalitck@asu.edu, mpeet@asu.edu, apachary@asu.edu }
}

\maketitle

\begin{abstract}

The immune response is a dynamic process by which the body determines whether an antigen is self or nonself. The state of this dynamic process is defined by the relative balance and population of inflammatory and regulatory actors which comprise this decision making process. The goal of immunotherapy as applied to, e.g. Rheumatoid Arthritis (RA), then, is to bias the immune state in favor of the regulatory actors  - thereby shutting down autoimmune pathways in the response. While there are several known approaches to immunotherapy, the effectiveness of the therapy will depend on how this intervention alters the evolution of this state. Unfortunately, this process is determined not only by the dynamics of the process, but the state of the system at the time of intervention - a state which is difficult if not impossible to determine prior to application of the therapy. {\color{black} To identify such states we consider a mouse model of RA (Collagen-Induced Arthritis (CIA)) immunotherapy}; collect high dimensional data on T cell markers and populations of mice after treatment with a recently developed immunotherapy for {\color{black} CIA}; and use feature selection algorithms in order to select a lower dimensional subset of this data which can be used to predict both the full set of T cell markers and populations, along with the efficacy of immunotherapy treatment.

\end{abstract}

\begin{IEEEkeywords} Immune State, Immunotherapy, Feature Selection, Rheumatoid Arthritis, Flow Cytometry
\end{IEEEkeywords}

\section{Introduction}\IEEEPARstart{W}{hile} a properly functioning immune system prevents illness by recognizing nonself antigens as foreign, a malfunctioning immune system can recognize self antigens as foreign, causing autoimmune diseases such as Rheumatoid Arthritis (RA).  In recent years immune therapies have been proposed that attempt to treat autoimmune diseases such as RA by shifting the relative balance between inflammatory and regulatory immune responses in favor of the regulatory populations.  For example, sustained delivery of chemokines~\cite{fisher2020situ,ratay2017treg}, cytokines~\cite{ratay2017tri} and small molecule inhibitors~\cite{acharya2017localized,ratay2017tri, jaggarapu2023orally} can modulate immune cell function (e.g. dendritic cells, T cells) in inflamed tissues to resolve RA and other autoimmune disease outcomes in pre-clinical animal models. However, the effect of the immunotherapy regimen is influenced by factors such as timing, dosage, and the current balance of inflammatory/regulatory response in the patient - thus making identification of effective treatment standards a challenging problem~\cite{mangal2021immunometabolism}.

For this reason, there is a growing need for an observable measure of immune system health which can be used for the prediction and prevention of RA and other autoimmune diseases~\cite{brodin2017human, davis2008prescription, germain2011human}. However, the question of identifying observables is complicated by our relative lack of understanding of how the immune system determines self vs nonself and the number of potential observables which have been identified as contributing to the function of the immune system. To clarify the problem at hand, we therefore propose two relatively uncontroversial theses.

{\color{black}
First, we presume that the immune system is governed by some un-modelled dynamical process wherein the relative populations of certain immunogenic and regulatory cells and molecules evolve over time and that the relative balance of these populations directly influences the establishment or elimination of autoimmune disease. Furthermore, we assume that this dynamic process is well-posed so that the inputs and initial states of the system uniquely determine the output (i.e. self-nonself). These unknown inputs and states are then potential observables.


Second, we presume that the problem of data-based modeling cannot be separated from the problem of identifying suitable observables. Specifically, if we knew which observables uniquely determined the output, then that knowledge would necessarily be based on some assumed physical model of the mechanism for producing that output. Thus, if the model is truly unknown, identification of observables must be included in the modeling process.


}



Given these assertions, we can propose three necessary components of any process for identification of observables with clinical predictive power. First, we require a method for modeling based on a given set of observables.  {\color{black} While such a model may be based on physical principles, the model may also be derived from data-based methods such as machine learning.} Second, we require a way to test suitability of the predictive model associated with any given set of observables. Specifically, this test of suitability may include predictive accuracy of the associated model, along with other metrics such as clinical feasibility and robustness to patient variation. Finally, we require a methodology for selection and rejection of observables in order to obtain a set of observables with maximal suitability as defined previously. In this paper, we consider each of these requirements: using experimental data and a variety of machine learning algorithms to generate models; defining an appropriate metric for suitability; and using feature selection algorithms to find a set of observables with maximal suitability.
 Once we have addressed these requirements, we apply the proposed methodology to data obtained from immunotherapy trials in an autoimmune mouse model of RA - arriving at a set of maximally suitable observables, which we define as the ``immune state''.
An outline of our approach to addressing these required subproblems is listed below. 

{\color{black}  For the first component we focus on machine learning algorithms for nonlinear regression. In the context of the immunotherapy, these regression algorithms map initial flow cytometry data to other observables such as outcome -- as measured by severity of inflammation (See Sec.~\ref{sec:comp_methods} for details).}

For the second {\color{black} component} we propose a dual metric for suitability of a given set of observables based partially on predictive power of the associated model. The first part of this metric is based on minimality (not prediction), wherein we impose a penalty based on the number of observables in the set (cardinality) in order to reduce experimental and clinical complexity. Second, in order to ensure that relevant immunological data is not lost, we also add a penalty based on the error of the associated model to predict observables from the data not included in the given set. Third, to measure efficacy of the prediction, we impose a penalty based on the error in prediction of {\color{black} CIA} severity - a quantity we refer to as the ``disease state''.

For the third {\color{black} component} we propose a variety of feature selection algorithms to determine the set of observables which are optimally suited using the suitability metric described above. We then report the results of applying the resulting algorithms to our dataset where we apply different weights to the three parts of the suitability metric and propose sets of maximally suitable observables for each case.  We define the optimal sets as "immune states" and analyze the immune cells that were selected by the feature selection algorithms in each case.

{\color{black} The rest of the paper is organized as follows. In Sec.~\ref{sec:data}, we define the dataset which will be used to generate observables (flow cytometry markers of predictive power). In Sec.~\ref{sec:comp_methods}, we define the computational and mathematical framework to be employed. This includes the learning algorithms used to produce the predictive models, a rigorous mathematical formulation of the feature selection problem defined in terms of a suitability metric, and a proposed wrapper algorithm for solving the feature selection problem. In Sec.~\ref{sec:results}, we apply the methods from Sec.~\ref{sec:comp_methods} to the dataset in Sec.~\ref{sec:data} using three possible suitability metrics defined as Minimal Disease State (MDS), Minimal Immune State (MIS), and a combined Minimal Immune and Disease State (MIDS). The results are summarized in Sec.~\ref{sec:discussion}.}

\section{Experimental Method and Associated Data}\label{sec:data}

{\color{black}The identification of observables for immune state described in this paper is based on a dataset generated from a series of experiments involving the use of biomaterials-based particles~\cite{mangal2020metabolite} containing metabolites to promote self tolerance in intermediate/late stage CIA in a DBA/1j mouse model. 
This study is well-suited to our hypothesis that immune state can serve to accurately predict the outcome of immunotherapy treatments and overall disease progression.
This section details the premise and execution of this study and the nature of the associated data collected.
}


\subsection{Fabrication of biomaterials-based particles}

{\color{black} Immunosuppressive poly aKG (paKG(PFK15+bc2) microparticles (MPs) have been developed to co-deliver the glycolytic inhibitor, PFK15, and the CIA-specific antigen, bc2 (bovine collagen type II), to mice with collagen-induced arthritis (CIA)~\cite{jaggarapu2023alpha, inamdar2023biomaterial, mangal2022short}. The underlying hypothesis revolves around the degradation of paKG MPs, which allows the delivery of bc2 to facilitate antigen presentation by dendritic cells (DCs), while simultaneously delivering PFK15 to attenuate glycolysis and CD86 expression in pro-inflammatory DCs. Furthermore, the intracellular release of PFK15 and aKG within DCs could collaboratively meet the cellular energy needs through the Krebs cycle, potentially curbing the energy requirements associated with pro-inflammatory glycolysis and fostering the generation of anti-inflammatory DCs. This orchestrated induction of anti-inflammatory DCs may consequently trigger suppressive antigen-specific T cell responses. This study underscores the potential of reprogramming DC metabolism, coupled with antigen presence, to instigate anti-inflammatory DC and T cell reactions, effectively alleviating arthritis symptoms in CIA mice. This innovative microparticle technology holds promise for addressing autoimmune diseases with a similar pathogenesis as rheumatoid arthritis~\cite{thumsi2023vaccines}.}

\subsection{Description of Experiment and Measurements}

\begin{figure}[t]%
\centering
\includegraphics[trim=10 0 300 20,clip,width=0.65\textwidth]{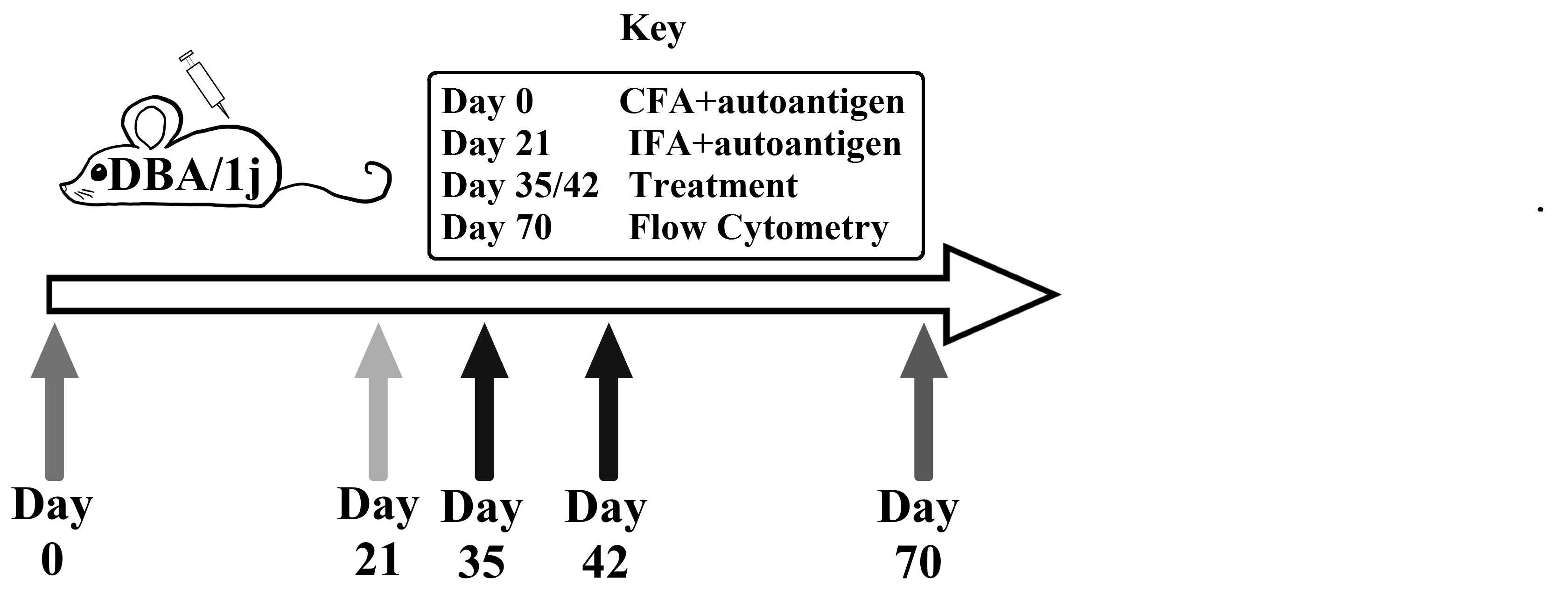}
 \caption{A graphical description of the experimental procedure of inducing and treating {\color{black} CIA} in mice. The first two steps induce {\color{black} CIA}, the next two steps is the application of the treatment and the final step is the data generation using flow cytometry.  CFA = complete Freund's adjuvant, IFA = incomplete Freund's adjuvant.} \label{fig:Experiment}
\end{figure}

Six to eight week old male DBA/1j mice were obtained from Jackson Laboratories and, after one to two weeks of mice acclimating to the experimental location, mice were induced with {\color{black} CIA}. In this experimental series, the particles were synthesized either with or without {\color{black} disease-inducing antigen} bc2 - a strategy designed to determine if the particles can generate antigen-specific anti-inflammatory response. 
The number of mice per group were determined using a statistical power of 80 percent and a significance level of alpha of 0.05. The arthritic scores were utilized to randomize mice into the control and treatment groups to assure that the overall average arthritic scores were comparable between each group. Researchers were aware of the group allocation throughout the study. An overview of the experimental procedure is provided in Fig.~\ref{fig:Experiment} and is further described in~\cite{mangal2021inhibition}. The chronology of the experiment is listed here in detail. 

{\bf Day 0 and 21:} {\color{black} CIA} was induced in mice to generate an autoimmune response for the development of severe polyarthritis. On day 35, the mice were divided into 3 groups, each receiving a distinct therapeutic regimen.

{\bf Group 0 {\color{black}(control)} - Days 35/42:}  The control group consists of 5 control mice, each receiving two subcutaneous injections of phosphate buffered saline (PBS) near the hind legs on days 35 and 42.

{\bf Group 1 {\color{black}(placebo)} - Days 35/42:}  Treatment group 1 consists of 5 mice. Each mouse receives two injections of 0.5 mg of biomaterials-based particles without embedded {\color{black}antigen} bc2 near the hind legs on days 35 and 42.

{\bf Group 2 {\color{black}(treatment)} - Days 35/42:}  Treatment group 2 consists of 8 mice. Each mouse receives two injections of 0.5 mg of biomaterials-based particles with embedded {\color{black}antigen} bc2 near the hind legs on days 35 and 42.

{\bf Measurements Taken on Days 62/70:}  The data collection used for model generation occurs exclusively on either day 62 or 70. Paw thickness measurements are used to determine arthritic scores for all mice and the end of study paw measurements were obtained either on day 62 or 70.
{\color{black} To quantify the range of variability and severity in paw swelling within the CIA model,  paw inflammation was evaluated on a scale from 0 to 6, where a score of 3 or higher indicated moderate-to-severe arthritis. The scores were determined by assessing the degree of swelling and/or redness in the rear left digits, with points assigned as follows: 0 (no swelling), 1 (mild), 2 (moderate), or 3 (severe). Similarly, the degree of swelling and/or redness in the rear left mid-paw was evaluated and assigned a corresponding point (0, 1, 2, or 3). This assessment was repeated for rear right digits and mid-paw. The cumulative points for the four assessments then determine the Disease Progression Score (DPS), as defined in Table~\ref{tab:scoring}. Scoring was carried out separately for the front and back paws.} The mice were euthanized by carbon dioxide asphyxiation according to the American Veterinary Medical Association (AVMA) guidelines and flow cytometry was performed on cells collected from the popliteal lymph node, cervical lymph node and spleen of each mouse on day 62 or 70.

\subsection{Observables Measured in the Dataset} 
{\color{black} The flow cytometry data obtained from this experiment was stained and gated to provide a robust set of features/(marker combinations)/observables. In Sec.~\ref{sec:comp_methods} we will define an algorithm for selection of the best lower-dimensional subset of these markers for predicting disease progression and representation of overall immune state. In this subsection,  we briefly describe the full selection of markers and (for motivation) list biological characterizations with which they are often associated. Specifically, we stained for: \textit{CD4} (T helper (Th) cell marker), \textit{CD8} (cytotoxic T (Tc) cell marker), \textit{Ki67} (proliferation), \textit{CD25} (activation), \textit{Foxp3} (regulatory Tcell transcription factor (TF)), \textit{Tbet+GATA3-RORyt-} (Th1/Tc1 TF), \textit{GATA3+Tbet-RORyt-} (Th2/Tc2 TF), \textit{RORyT+Tbet-GATA3-} (Th17/Tc17 TF), \textit{CD44} (effector memory marker), \textit{CD62L} (na\"ive T cells and central memory T cells), and a tetramer (I-A(q)	bovine collagen \uppercase\expandafter{\romannumeral 2\relax} 271-285, GEPGIAGFKGEQGPK peptide) that is specific to the  disease-inducing antigen. For notational convenience, we use \textit{GATA3+}, \textit{RORyt+} and \textit{Tbet+} as an indicator of \textit{GATA3+Tbet-RORyt-}, \textit{RORyT+Tbet-GATA3-} and \textit{Tbet+GATA3-RORyt-}, respectively. The flow cytometry data analysis was performed by comparison of Forward scatter (FSC) vs. side scatter (SSC), then single cell (SSC-A vs. SSC-H), followed by live dead stain, followed by CD4 vs. CD8, followed by individual sub-populations of the T cells~\cite{mangal2021inhibition}. Based on this staining, we identified 41 different combinations of markers which might be used to classify the phenotype of a T cell and determined the percentage of either CD4 or CD8 T cells presenting the associated combination of markers.  }

\begin{table}[t]
    \centering
\begin{tabular}{|c|c|}
\hline
Total points for the mouse & Disease progression score \\
\hline
     0     &  0  \\
     1     &  1  \\
     2     &  2  \\
     3-4   &  3  \\
     5-7   &  4  \\
     8-10  &  5  \\
     11-12 &  6  \\
     \hline
\end{tabular}
    \caption{\color{black}Scoring strategy for mice. The degree of redness and swelling has been measured for rear right digits, rear left digits, mid-paw and rear left mid-paw based on points out of 3, where  0 (no
swelling), 1 (mild), 2 (moderate), or 3 (severe). The total points for the mouse is a cumulative points for paws and digits, that has been mapped to a Disease Progression Score (DPS).}
    \label{tab:scoring}
\end{table}


\subsection{Summary of Associated Dataset}

The data consists of 84 samples, based on 18 mice, each sample is associated with a mouse and sample location. There were no exclusions of mice, experimental units or data points. All samples are taken on day 62/70, and each sample consists of 43 features and one label. The first two features of each sample indicate group number (0-2) and sample location (1-3). The remaining 41 features defining the percentage (0-100) of the CD4/CD8 population exhibiting the associated combination of markers. {\color{black}The label for each sample is a Disease Prograssion Score (DPS) (0-6).}

{\color{black}
This dataset has three crucial properties. First, measured features represent the significant part of the immune system, since we identify the wide range of  measured T cell markers and populations to reflect the majority of aspects of autoimmune diseases, such as self-reactivity, memory, activation and proliferation.  Second, the collected data includes a full list of possible stages of disease, so we include mice with and without CIA in our experiment. Finally, the dataset has minimimal distortion effects of non-measurable observables at measurements days, for example the distorting effects of the treatment must wear off and measured features then accurately reflect the immune state. In this case, measurements were taken more than two weeks after the treatment.   }

Based on this data we are ready to propose several methods of machine learning to construct predictive models which use subsets of the features to predict both the label (disease progression) and remaining features. For generating these models, all features are scaled to the interval $[0,1]$.

\section{Computational Methods for identification of Observables}
\label{sec:comp_methods}
{\color{black} In this section, we describe a general mathematical and computational approach for generating observables using machine learning and feature selection. We start with a description of the several candidate algorithms to be used for generating predictive models (Subsec.~\ref{subsec:MachineLearning}). Next, in Subsec.~\ref{subsec:Suitability} we propose a mathematical formulation of the feature selection problem, using a metric for suitability of a set of observables in terms of the predictive model generated by those observables. Then, in Subsec.~\ref{subsec:FS} we describe the}
{\color{black}proposed feature selection algorithm along with alternatives to be used for comparison.}
\subsection{Algorithms for Predictive Model Generation}
\label{subsec:MachineLearning}
To identify clinically significant observables, we will use a metric of suitability combined with a feature selection algorithm to determine which observables have the most predictive power. However, the use of such feature selection algorithms requires a procedure for using a subset of the features to predict both the remaining features and the label.

Suppose we are given a dataset of $m$ samples, wherein each sample $\{x_i,y_i\}$ defines a set of features $\{x_i \in \R^n \}_{i=1}^{m}$  and an associated label $\{y_i \in \R \}_{i=1}^{m}$. The regression problem, then, is to find a predictive model, $f:\R^n\rightarrow \R$ which minimizes the predictive errors $f(x_i)-y_i$ in an appropriately defined metric. However, this metric and the resulting optimization problems vary significantly between algorithms. We next define several state-of-the-art machine learning algorithms which will be combined with feature selection algorithms to determine features with the most predictive power. Finally, we note that in the context of feature selection algorithms, when only a subset of the available features are used, the remaining ``discarded'' features become labels.



Before beginning, we note that the choice and tuning of ML algorithms is something more of an art than a science. Specifically, we want to avoid overfitting the training data - thus allowing our predictive models to perform well on unlabelled data. To this end, each of the ML algorithms we define has an associated set of ``regularization parameters'' which should be selected through some ad hoc process. These tuning parameters will then affect how well the resulting predictive model will generalize to unlabeled data. In each case, therefore, we specify these parameters but do not yet define how they are selected.

In each case below, we assume the data set contains $m$ samples, $\{x_i,y_i\}_{i=1}^m$, each with $n$ features, $x_i\in \R^n$ and a label $y_i \in \R$.

{\bf Regularized Linear Regression (LR)}:
The regularized linear regression algorithm returns a predictive model $y=f(x)=w^Tx+b$, where $w$ solves the following optimization problem.
\vspace{-4mm}
\[ \min_{w \in \R^n} ~~ \sum_{i=1}^m (y_i - w^T x_i-b)^2 + \alpha_2||w||^2 + \alpha_1 ||w||.\]
In this case, $\alpha_1 \geq 0$ and $\alpha_2 \geq 0$ are the regularization parameters. Linear regression has the advantage of low computational complexity. However, the resulting predictor is linear and if the underlying physical process is nonlinear, accuracy of the predictive model will be poor.

{\bf $\epsilon$-loss Support Vector Regression (SVR)}:
The support vector regression problem uses a predictive model of the form $f(x) = \sum_{i=1}^m \alpha_i k(x,x_i)$ where $\alpha \in \R^m$ is the decision variable and $k$ is a user selected positive kernel function. The objective function being minimized includes $\sum_i |f(x_i) - y_i| $ for any $i$ such that $|f(x_i) - y_i| \geq \epsilon$, where $\epsilon$ is a tuning parameter. In addition, there is a regularization parameter, $C$ where regularization increases as $C$ decreases. SVR can generate accurate nonlinear predictive models for appropriate choice of $k$. However, the selection of the kernel heavily influences the resulting accuracy and this process of selection is difficult to automate.

{\bf Kernel Learning (PMKL)}:
Kernel learning algorithms improve on the SVR problem by automating the search for a kernel function. Note we consider the class of kernel learning algorithms to include Deep Learning (although the search problem in this case is non-convex). These approaches are limited, however, by the class of kernels over which they are able to search. The class of Tessellated Kernel functions have been shown in~\cite{jmlr} to have the properties of universality, density, and tractability - meaning the resulting algorithms are rather accurate and generalize well to new data. Specifically, the PMKL algorithm for optimizing TK kernel functions was shown in~\cite{colbert2020new} to be more robust than other tested ML algorithms (including multi-layer neural networks) - at the cost of some additional computational complexity.  The regularization parameters in this case are the $\epsilon$ and $C$ as defined above for SVR.

{\bf Decision Tree Algorithms}:
Decision trees are composed of a series of conditional statements that branch in a ``tree'' like manner.  We say the ``depth'' of a decision tree is how many conditional statements appear in a branch before leading to a label denoted the ``leaf''.  Both the depth of the decision trees and the maximum number of leaves are regularization parameters that can be modified by the user.  Decision trees are often weak predictors alone and in this paper we use ensemble (random forest) or boosting (boosted trees) methods to increase predictive performance. These algorithms are defined as follows.
\begin{itemize}
\item \textbf{Random Forest:}
The random forest algorithm is an ensemble machine learning method based on a combination of decision trees.  Ensemble methods use a combination of predictive models (trees) that individually have poor generalization but when used in combination can have significantly improved predictions.  The number of decision trees combined in the random forest algorithm can be used as a regularization parameter.
\item \textbf{Boosted Trees:}
Gradient boosting is another machine learning method also based on a combination of decision trees.  In the boosted algorithm trees are added to the predictive model sequentially, and each additional tree is fit to the current residuals of the model.  A ``learning rate'' is a weight applied to the addition of each decision tree, and is often used as a regularization parameter.  Small learning rates tend to improve the generalization of the predictive models.
\end{itemize}

Next we will focus on a metric we may use to identify the observables which are most suitable to the task of predicting self vs nonself determination in autoimmune disease.

\subsection{Quantifying Suitability of a Given Set of Observables} \label{subsec:Suitability}

To identify a set of observables for predicting self vs nonself determination we rigorously define a metric for suitability in order to select the observables which lead to superior predictive models.

First, for the sake of generality, we define the algorithm, $OPT$, which we use as a placeholder for the machine learning algorithms described previously. For a given dataset $\{x_i,y_i\}_{i=1}^m \subset \R^w \times \R^q$,  $OPT(\{x_i,y_i\}_{i=1}^m)$, returns a predictive function, $f=\text{arg}\; OPT(\{x_i,y_i\}_{i=1}^m) $, where $f: \R^w \rightarrow \R^q$.

Next, given a possible set of feature indices $F:=\{1,\cdots,n\}$, we define the set of partitions of $F$ as $\mcl P(F)$, and the set of all possible partitions of $F$ of length $w\le n$ as follows.
\[
B_w := \{ v \in \N^w \; | \; v \in \mcl P(F) \}
\]
For a given selection of features, $b \in B_w$, we denote the associated projection $P_b:\R^n \rightarrow \R^w$ so that $(P_b(x))_{i}=x_{b_i}$ for $x \in \R^n$ and $i=1,\cdots, w$.

To define a metric of suitability we consider three cost/penalty functions, $M_1, M_2,$ and, $L$. The function $L$ is a function of the cardinality of the number of features selected, $L(|b|_C)$. The costs $M_1$ and $M_2$, however, measure how well the selection of features can be used to predict the disease state and the remaining features respectively.  To accurately evaluate the performance of the predictor a partition of the data must be withheld from the training algorithm, $OPT$, and used solely for the purpose of testing the performance.  For a given set of data, these metrics will vary depending on which data points are used for training $OPT$ and which are used to evaluate its performance. To explicitly account for the effect of choice in partitioning of data samples, we now define the set of samples $S:=\{1,\cdots,m\}$, and the set of partitions of $S$ as $\mcl P(S)$. As for features, we denote the set of sample partitions of length $r$ as
\[
S_r := \{ v \in \N^r \; | \; v \in \mcl P(S) \}
\]
and for a given selection of samples, $g \in S_r$, we denote the associated projected data set as $\mcl P_g(X):=\{x_i \in X,\; i \in g\}$.

Therefore, the costs $M_1$ and $M_2$ are a function of the feature partition, $b$, the training partition, $g \in S_r \in \mcl P(S)$ and the associated test partition, $h :=S/g \in S_{m-r}$, so that $M_1(b,g)$ and $M_2(b,g)$ are the Root Mean Square Error (MSE) of predicting the test partition. Specifically, let $R(f,x,y) = \sqrt{\frac{1}{m-r}\sum_{i \in S/g} \left| f(x_i) - y_i \right|^2_2}$ and we have
\begin{align}
&M_1(b,g) = R(f_{b,g},P_{b}(x),y) 
\nonumber  \\
&M_2(b,g) = \sum_{j \in F/b} R(d_{b,g}^{(j)},P_{b}(x),P_{j}(x)) \nonumber \\ 
&f_{b,g} = \text{arg}OPT(\{P_b(x_{g_i}),y_{g_i}\}_{i=1}^r) 
\nonumber \\
&d_{b,g}^{(j)} = \text{arg}OPT(\{P_b(x_{g_i}), P_{j}(x_{g_i})\}_{i=1}^r)) \nonumber
\end{align}

In the ideal case, we would average these costs over all possible partitions of the data set to give an estimate of the predictive power of $b \in B_w$. However, such an approach would result in very large computational overhead. Therefore, we use the $k$-fold cross validation approach, wherein we divide the samples into $k$ training partitions of size $\frac{m(k-1)}{k}$, which we label as $g(i) \in S_{\frac{m(k-1)}{k}}$ for $i=1,\cdots,k$. Then the average cost of the feature partition $b$ over the $k$ sample partitions is
\vspace{-2mm}
\[
J(b)=\frac{1}{k} \sum_{i=1}^k J'(b,g(i))\vspace{-2mm}.
\]
\vspace{-2mm}
where
\begin{equation}
J'(b,g)  :=  \beta_1 M_1(b,g) + \beta_2 M_2(b,g) + L(|b|_C)\label{eqn:suitability}
\end{equation}
and where $\beta_1,\beta_2 \geq 0$ are given weights. {\color{black} In Sec.~\ref{sec:results}, we use specific valued of $\beta_1,\beta_2$ and $L$ to define three possible suitability metrics relevant to identification of immune state. These will be referred to as Minimal Disease State (MDS), Minimal Immune State (MIS), and the combined Minimal Immune and Disease State.

In the following Subsection, we propose a feature selection algorithm which can be used to select observables which optimize suitability metrics as defined in Eqn.~\eqref{eqn:suitability}.}

\subsection{Feature Selection Algorithms} \label{subsec:FS}
{\color{black} We have now defined the metric of suitability as a function of the partition, $b \in B_w$. Using this metric, the Feature Selection (FS) problem is defined as the following combinatoric optimization problem.
\begin{align} \label{opt:FSProblem}
    \min_{b \in B_w, w \in \N} ~~ & ~~ J(b)
\end{align}

Because optimization problems of this form are combinatorial, FS problem is considered to be NP-hard~\cite{Chandrashekar2014ASO}. As a consequence, most existing FS algorithms are either heuristic, in that they are not guaranteed to converge to a globally optimal solution, or solve variations of this problem which may or may not yield reasonable values for Problem~\eqref{opt:FSProblem}.}

Nonetheless, several techniques have been proposed that enjoy relative accuracy and computational efficiency.  We focus first on FS methods designed specifically for problems of the same form as Optimization Problem~\eqref{opt:FSProblem}, then consider two other FS approaches that do not directly try to solve the optimization problem of interest but provide a comparison to the direct method.


\subsubsection*{Proposed Wrapper Method and Implementations}\label{subsubsec:wrapper}
We first define the algorithm (a wrapper method) which will be used and then provide additional details on the various ML algorithms which are combined with this wrapper to solve Problem~\eqref{opt:FSProblem}.

The most common wrapper methods are Sequential Feature Selection (SFS) algorithms~\cite{Chandrashekar2014ASO}.  SFS algorithms begin with an empty (or full) set of features and sequentially add (or remove) the highest value (or cost) feature until the set of features is a certain size or meets a performance metric.

The SFS algorithm used in this paper is as described in~\cite{dash1997feature}.  This SFS algorithm begins with $b := \emptyset$, and iteratively selects a locally optimal feature (with respect to the objective function of Optimization Problem~\eqref{opt:FSProblem}) at each step.

Clearly, the effectiveness of Feature Selection depends on the ML algorithm ($OPT$) used to generate the predictive model. Therefore, in the Results Section, we test all the machine learning algorithms proposed herein. Unfortunately, the accuracy of the predictive model is influenced by user-selected parameters within the algorithm. For reproducibility, we list here the selections for these parameter values.

\noindent \textbf{Linear Regression:} We test all 16 combinations of $\alpha_1 \in[0, 0.1, 1, 5] $ and $\alpha_2 \in[0, 0.1, 1, 5]$  and the data from choice yielding highest suitability ($J$) is listed in Tables~\ref{tab:results_MDS}-\ref{tab:results_MOS}.\\
  \noindent  {\bf PMKL:} We use the default TK kernel parameters and test $\epsilon = .005$, and $C \in [.01,.1,.3,.5,1]$ and the data from the choice yielding highest suitability ($J$) is listed in Tables~\ref{tab:results_MDS}-\ref{tab:results_MOS}.\\
\noindent {\bf SVR:} We test all combinations of $\epsilon = .1$, $C \in [1,5,10]$ and 3 kernel functions (linear, RBF, or 3rd degree polynomial) and the data from choice yielding highest suitability ($J$) is listed in Tables~\ref{tab:results_MDS}-\ref{tab:results_MOS}.  For the RBF kernel the features are normalized by their variance and a bandwidth of $\frac{1}{n}$ is selected.  \\
  \noindent  {\textbf{Random Forest}} We test 9 combinations of number of trees ($\text{n}_{\text{trees}} \in [50, 100, 150]$) and the maximum tree depth of ($\text{max}_{\text{depth}} \in [5, 10, 20] $) and the data from choice yielding highest suitability ($J$) is listed in Table~\ref{tab:results_MDS}-\ref{tab:results_MOS}. \\
\noindent \textbf{Boosted Trees} We test 15 combinations of number of trees ($\text{n}_{\text{trees}} \in [50, 100, 150, 250]$) and  learning rate ($\text{LR} \in [0.01, 0.1, 0.5]$) and the data from choice yielding highest suitability ($J$) is listed in Table~\ref{tab:results_MDS}-\ref{tab:results_MOS}.


\subsubsection*{Suitability of Filter and Embedded Methods} \label{subsubsec:filter_embedded}
Alternative feature selection algorithms will be used as a baseline by which we may compare the wrapper method.  We use three filter methods and one embedded method in the analysis.

Given a set of data, \textit{filter methods} use a rating function to rank each features relative ``importance''.  After the features have been ranked, the user may select $w$ features to be kept and the remaining features will be discarded. The rating functions used to generate the data in Tables~\ref{tab:results_MDS}-\ref{tab:results_MOS} are as follows.

 {\bf Mutual Information (MI)}: The Mutual Information criteria~\cite{Battiti1994UsingMI} is a statistical function of two random variables that describes the amount of information contained in one random variable relative to the other.

 {\bf Analysis of Variance (ANOVA)}: The ANOVA method ~\cite{kumar2015feature} is a commonly used method for analyzing variable dependencies. The \textit{F-test} is used to estimate the features importance.

 {\bf Principle component analysis (PCA)}: This method approximates the data with linear manifolds~\cite{Song2010FeatureSU}. The main methods used to perform PCA are based on the singular value decomposition and diagonalization of the correlation matrix. We calculate the importance based on the first 3 eigenvectors.

In all cases, once a set of features has been selected, suitability ($J$) is determined using each of the ML algorithms and we report the minimum of these values.
\begin{figure*}[!t]%
 \includegraphics[width=.98\textwidth]{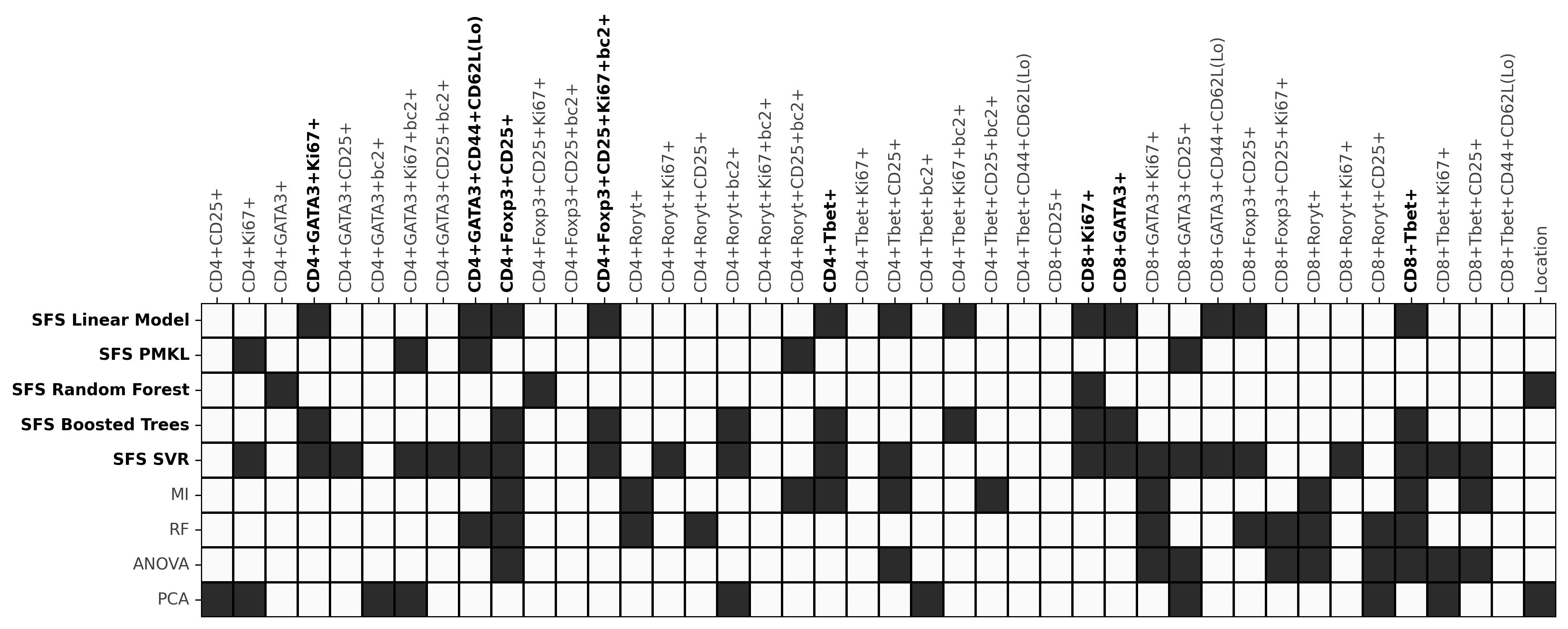}
	\caption{\color{black}Observables as selected by SFS with each of 5 ML algorithms  using the MDS metric for suitability. For comparison, we also include the observables selected by the 4 wrapper methods defined in Sec.~\ref{sec:comp_methods}. The methods are ordered from top to bottom as determined by the metric for suitability of the selected observables as defined in Subsec.~\ref{subsec:Suitability} and listed in Table~\ref{tab:results_MDS}. Evaluation of suitability for wrapper methods is as described in~\ref{subsec:FS}. The SFS methods and the features most commonly selected by those methods are bolded.} \label{fig:dotPlotMDS}
\end{figure*}
\textit{Embedded FS methods} attempt to embed the process of feature selection directly into the model generation process - typically adding a cost for inclusion of a particular feature in the model.  These methods have been used in the gene expression domains as in~\cite{Guyon2004GeneSF} and have been successfully applied to mass spectrometry analysis in~\cite{Zhang2006RecursiveSF, Jong2004FeatureSI, Prados2004MiningMS}.   For this analysis, only a single embedded method was considered.

{\bf Mean Decrease in Impurity (RF)}: The Gini Importance or Mean Decrease in Impurity~\cite{Breiman1984ClassificationAR} is an embedded method for the Random forest algorithm. It calculates the importance of features as the mean of the number of splits (over all trees) that include this feature, weighted by the probability of reaching this node.

\subsubsection*{Performance Metrics}
 To show that the results of Optimization Problem~\eqref{opt:FSProblem} as applied to MDS, MIS and MIDS are consistent with other learning metrics~\cite{Borchani2015ASO}, we also include data on these metrics for the chosen selection of features and associated predictor. These metrics are defined as follows. Let $y$ to be the vector of labels (measured non-selected features and the disease state) associated with features $x$. Let $\hat y$ be the predicted labels as generated by the predictor when applied to features $x$. Let $\bar{y}$ and $\bar{\hat{y}}$ be the average values of $y$ and $\hat{y}$. Then we have the following.\\
\textbf{The Correlation Coefficient and relative Root Mean Squared Error} (CC and rRMSE):
\begin{align}
&\text{CC} = \frac{\sum_{i = 1}^{N} (y_i - \bar{y}_i)(\hat{y}_i - \bar{\hat{y}}_i)}{\sqrt{\sum_{i = 1}^{N} (y_i - \bar{y}_i)^2 \sum_{i = 1}^{N} (\hat{y}_i - \bar{\hat{y}}_i)^2}} \label{eqn:CC}\\
& \text{rRMSE} =  \sqrt{\frac{\sum_{i = 1}^{N}(y_i - \hat{y}_i)^2}{\sum_{i = 1}^{N}(y_i - \bar{y}_i)^2}} \label{eqn:rRMSE}
\end{align}
\textbf{Mean Absolute Error and relative Mean Absolute Error} (MAE and rMAE):
\begin{align} \label{eqn:MAE}
\text{MAE} =  \frac{1}{ N }\sum_{i = 1}^{N}|y_i - \hat{y}_i|; \;\;\; \text{rMAE} =  \frac{\sum_{i = 1}^{N}|y_i - \hat{y}_i|}{\sum_{i = 1}^{N}|y_i - \bar{y}_i|}
\end{align}


\section{Results: Identification of Observables from a CIA Dataset}\label{sec:results}
{\color{black} In this section, we apply the methods defined in Sec.~\ref{sec:comp_methods} to the data described in Sec.~\ref{sec:data} to obtain three possible sets of observables (immune state) corresponding to different suitability metrics (as defined in Subsec.~\ref{subsec:Suitability}).  Specifically, these three immune states are lower dimensional subsets of the data which can be used to either predict the progression of   CIA, reconstruct the full} 
{\color{black}set of T cell markers and populations, or perform both tasks simultaneously.

All results obtained in this section were obtained using implementations of the ML regression algorithms either from}
{\color{black}scikit-learn 0.22.1 or PMKL v1~\cite{colbert2020convex}. Computation was performed on an Intel i7-5960X CPU with 128Gb of RAM. Details of this implementation, including all experimental data, FS wrapper, and and codes for regression have been made publically available and can be found at~\cite{githubMS}.}

\subsection{Suitability metrics for the CIA dataset}\label{subsec:suitability_metrics}

 {\color{black}  In this subsection, we define three metrics of suitability used for selecting observables.  First, we consider the feature selection problem for Minimal Disease State (MDS) using the definition of suitability in Subsec.~\ref{subsec:Suitability} with  $\beta_1 = 1$   and $\beta_2 = L(w) = 0$.  In this case, suitability is defined only in terms of accuracy of the prediction of the Disease Progression Score (DPS).    Results for MDS suitability are given in Subsec.~\ref{subsec:MDS}. Second, we define the feature selection problem for Minimal Immune State (MIS) using $\beta_1 = 0$ and $\beta_2 = 1$ and $L(w)=\begin{cases}
    0 & \text{for } w \leq 10 \\
    \infty & \text{for } w > 10.
  \end{cases}$ In this case, we ignore the DPS and restrict our definition of state to ten observables (flow cytometry markers), defining suitability as the ability to predict all markers not included in our chosen set of 10 observables. Results for MIS suitability are given in Subsec.~\ref{subsec:MIS}. Finally, for Minimal Immune and Disease State (MIDS), we let $\beta_1 = \beta_2 = 1$ and $L(w)$ as defined for MIS. In this case, we restrict our definition of state to 10 observables and define suitability as the ability to predict a weighted combination of the DPS and all markers not included in the chosen set of 10 observables. Results for MIDS suitability are given in Subsec.~\ref{subsec:MIDS}.}

\begin{figure*}[!t]
 \includegraphics[width=.98\textwidth]{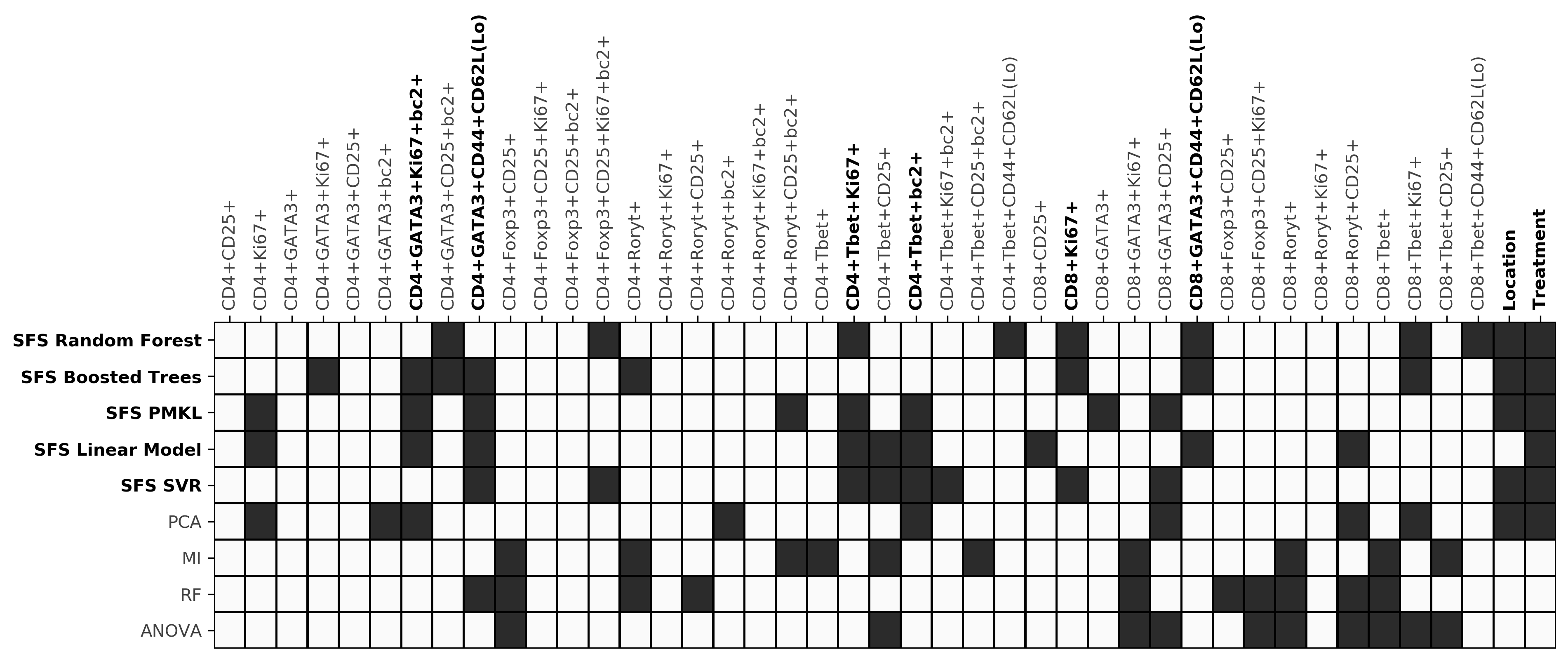}
	\caption{\color{black}Observables as selected by SFS with each of 5 ML algorithms  using the MIS metric for suitability. For comparison, we also include the observables selected by the 4 wrapper methods defined in Sec.~\ref{sec:comp_methods}. The methods are ordered from top to bottom as determined by the metric for suitability of the selected observables as defined in Subsec.~\ref{subsec:Suitability} and listed in Table~\ref{tab:results_MIS}. Evaluation of suitability for wrapper methods is as described in~\ref{subsec:FS}. The SFS methods and the features most commonly selected by those methods are bolded.} \label{fig:dotPlotMIS}
\end{figure*}
\begin{table}[h]
\centering
  \begin{tabular}{ |c|ccccc|}
    \hline
    {\bf Algorithm} & J & MAE & rRMSE & rMAE & CC \\
    \hline
Linear Model	&0.32 	&0.27 	&0.78 	&0.72 	&0.55 \\	
PMKL	&0.34 	&0.29 	&0.84 	&0.84 	&0.51 \\
Random Forest	&0.35 	&0.29 	&0.83 	&0.87 	&0.53 \\
Boosted Trees	&0.36 	&0.31 	&0.89 	&0.88 	&0.49 \\
SVR	&0.37 	&0.31 	&0.90 	&0.93 	&0.40 \\
MI  	&0.38 	&0.34 	&0.99 	&0.99 	&0.28 \\
RF &0.38 	&0.34 	&0.98 	&1.02 	&0.30 \\
ANOVA 	&0.38 	&0.35 	&1.02 	&1.01 	&0.28 \\
PCA 	&0.40 	&0.38 	&1.11 	&1.10 	&0.28 \\
    \hline
\end{tabular}
\newline
  \caption{\color{black}Comparison of metrics for the observables selected from the 5 proposed feature selection algorithms using the MDS suitability metric. The 4 wrapper methods described in Subsec.~\ref{subsec:FS} are included for comparison. The order of the methods are determined using the suitability metric (J). The other metrics (CC,rMSE, MAE, and rMAE) are as defined in Subsec.~\ref{subsec:FS}.}
 \centering
 \label{tab:results_MDS}
\end{table}
\subsection{The best features for Predicting Disease Progression (MDS)}\label{subsec:MDS}

{\color{black} First, we consider selecting observables (markers) which optimize suitability with respect to Minimal Disease State (MDS) as defined in Subsec.~\ref{subsec:suitability_metrics}. These are observables which are best at  predicting the disease progression score (DPS).

{\bf Performance of FS Algorithms}:
In Table~\ref{tab:results_MDS}  we rank the proposed feature selection algorithms by performance with respect to the MDS suitability metric, $J$ (as defined in Optimization Problem~\eqref{opt:FSProblem}). For comparison, we include other metrics of fit (as defined in Eqs.~\eqref{eqn:CC}-\eqref{eqn:MAE}), and alternative filter-based feature selection algorithms.

The results indicate that Sequential Forward Selection (SFS) based algorithms performed significantly better than embedded and filter methods with respect to all metrics.
Interestingly, although no weight or limit was placed on the number of features selected, the SFS Random Forest and the SFS TK both selected relatively few features (4 and 5, respectively) -- less than half as many features as the 12 average features selected by  other methods. This indicates that use of the remaining \textit{increased} error in the testing set -- meaning that} {\color{black}other observables are likely redundant or unreliable indicators
of disease progression.}

{\bf Most Important Features Using the SFS Algorithms}:
In Fig.~\ref{fig:dotPlotMDS} we show the observables that were selected by each of the proposed algorithms.  If we consider only the top performing algorithms (the SFS based algorithms) and the markers specific to helper and regulatory cells, then counting the number of times a feature was selected by the SFS algorithms, the following features were chosen by at least three of the algorithms.
\begin{enumerate}
\item[(1)] \textit{CD4+GATA3+CD44+CD62L(Lo)} (3 times)
\item[(2)] \textit{CD4+GATA3+Ki67+} (3 times)
\item[(3)] \textit{CD4+Foxp3+CD25+} (3 times)
\item[(4)] \textit{CD4+Foxp3+CD25+Ki67+{\color{black}bc2+}} (3 times)
\item[(5)] \textit{CD4+Tbet+} (3 times)
\end{enumerate}

Among the cytotoxic cells, the algorithms were most consistent, with all five of the algorithms selecting one feature in common.
\begin{enumerate}
	\item[(6)] \textit{CD8+Ki67+} (4 times)
	\item[(7)] \textit{CD8+GATA3+} (3 times)
	\item[(8)] \textit{CD8+Tbet+} (3 times)
\end{enumerate}

This group of cells consists of cytotoxic (6,7,8), Th memory (1), Th (2,5), and {\color{black}\textit{CD4+CD25+Foxp3+} regulatory} T cell sub-populations (3,4).  The \textit{location} feature (origin of the tested cells), was selected only once by an SFS based algorithm.  In this case we do not include the treatment as a possible feature, since we are primarily interested in the prediction of the disease state using sub-populations of T cells as opposed to the already known correlation between treatment and disease state. In the next two cases (MIS and MIDS)  treatment is considered a feature.

\begin{table}[h]
\centering
  \begin{tabular}{|c|ccccc|}
    \hline
    {\bf Algorithm} & J & MAE & rRMSE & rMAE & CC \\
    \hline
Random Forest	        &0.11 	&0.08 	&0.37 	&0.29 	&0.89 \\
Boosted Trees	&0.12 	&0.08 	&0.38 	&0.31 	&0.88 \\
PMKL	&0.12 	&0.09 	&0.37 	&0.22 	&0.89 \\
Linear Model	&0.13 	&0.09 	&0.44 	&0.31 	&0.87 \\
SVR	&0.13 	&0.09 	&0.46 	&0.32 	&0.85 \\
PCA 	&0.15 	&0.12 	&0.58 	&0.49 	&0.76 \\
MI &0.17 	&0.13 	&0.61 	&0.53 	&0.74 \\
RF	&0.17 	&0.13 	&0.59 	&0.49 	&0.75 \\
ANOVA	&0.18 	&0.14 	&0.67 	&0.60 	&0.71 \\ \hline
\end{tabular}
\newline
  \caption{\color{black}Comparison of metrics for the observables selected from the 5 proposed feature selection algorithms using the MIS suitability metric. The 4 wrapper methods described in Subsec.~\ref{subsec:FS} are included for comparison. The order of the methods are determined using the suitability metric (J). The other metrics (CC,rMSE, MAE, and rMAE) are as defined in Subsec.~\ref{subsec:FS}.}
 \label{tab:results_MIS}

\end{table}

\subsection{The best features for Reconstructing Discarded Features (MIS)}\label{subsec:MIS}


{\color{black} Next, we consider selecting observables (markers) which optimize suitability with respect to Minimal Immune State (MIS). These observables are optimal for predicting all markers not included in our chosen set of 10 observables.

{\bf Performance of FS Algorithms}:
In Table~\ref{tab:results_MIS}  we rank feature selection methods by performance with respect to the MIS suitability, $J$. For comparison, we also include other metrics and filter-based methods as defined in Subsec.~\ref{subsec:FS}.

The results show that Sequential Forward Selection (SFS) based algorithms demonstrated the best performance. Especially, SFS Random Forest and SFS Boosted Trees are the best methods with respect to MIS suitability metric.  Note, that all algorithms selected the maximum number of 10 features.}
\begin{figure*}[t!]%
 \includegraphics[width=.98\textwidth]{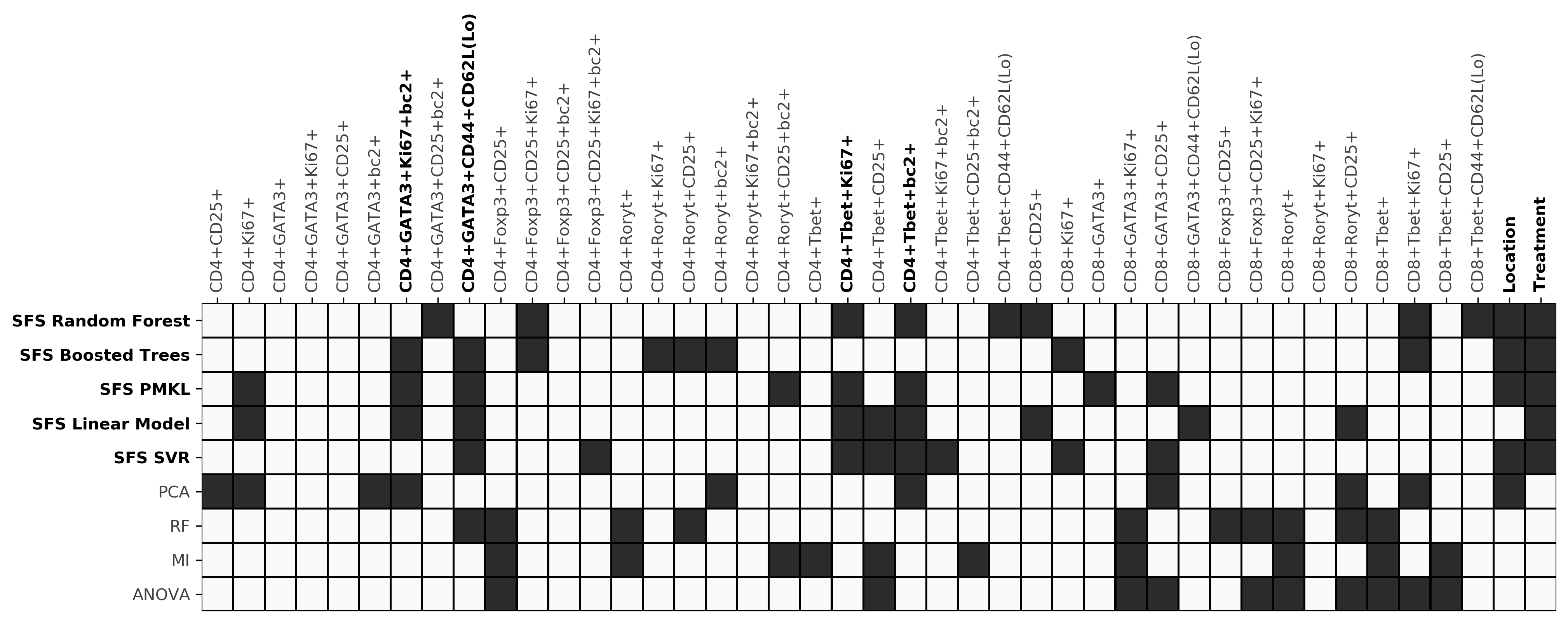}
\caption{\color{black}Observables as selected by SFS with each of 5 ML algorithms  using the MIDS metric for suitability. For comparison, we also include the observables selected by the 4 wrapper methods defined in Sec.~\ref{sec:comp_methods}. The methods are ordered from top to bottom as determined by the metric for suitability of the selected observables as defined in Subsec.~\ref{subsec:Suitability} and listed in Table~\ref{tab:results_MOS}. Evaluation of suitability for wrapper methods is as described in~\ref{subsec:FS}. The SFS methods and the features most commonly selected by those methods are bolded.} \label{fig:dotplotMOS}
\end{figure*}

{\bf Most Important Features Using the SFS Algorithms}:
In Fig.~\ref{fig:dotPlotMIS} we show the features that were selected by each of the proposed algorithms. Unlike in the previous subsection, there was less of an agreement among the high-performing SFS algorithms as to the most significant features.  For MIS only 6 different features were selected by at least three algorithms.  First, if we consider markers specific to helper and regulatory cells, and counting the number of times a feature was selected by the SFS methods (each method selected 10 features), the following features were each chosen by at least 3 algorithms.
\begin{enumerate}
\item[(1)] \textit{CD4+GATA3+CD44+CD62L(Lo)} (4 times)
\item[(2)] \textit{CD4+Tbet+Ki67+} (4 times)
\item[(3)] \textit{CD4+GATA3+Ki67+{\color{black}bc2+}} (3 times)
\item[(4)] \textit{CD4+Tbet+{\color{black}bc2+}} (3 times)
\end{enumerate}
We note that two of the selected features are {\color{black} bc2} specific as opposed to the single {\color{black} bc2} specific feature selected for cells in MDS.

Among the cytotoxic cells, the algorithms were less consistent, with only three of the algorithms selecting similar sub-populations.
\begin{enumerate}
\item[(5)] \textit{CD8+Ki67} (3 times)
\item[(6)] \textit{CD8+GATA3+CD44+CD62L(Lo)} (3 times)
\end{enumerate}

We note that the central memory T cells {\color{black}(\textit{CD44+CD62L(Lo)})} appear in both the helper/regulatory populations and the cytotoxic cell populations.  In this case, data-rich biomarkers (those containing multiple markers), were selected slightly more often when compared to MDS.  The average number of markers in the selected features is 3.33 in this case compared to 2.875 in the MDS case.
Of particular note is the fact that the \textit{location} feature (origin of the tested cells) and the \textit{treatment} feature (which treatment was applied) were both selected by almost every algorithm.

\begin{table}[h]

\centering
  \begin{tabular}{ |c|ccccc| }
    \hline
    {\bf Algorithm}   & J & MAE & rRMSE & rMAE & CC\\
    \hline
Random Forest	&0.11 	&0.09 	&0.38 	&0.29 	&0.90 	\\	
Boosted Trees	&0.12 	&0.09 	&0.39 	&0.31 	&0.88 	\\
PMKL	&0.12 	&0.09 	&0.38 	&0.23 	&0.89 	\\
Linear Model	&0.13 	&0.10 	&0.45 	&0.31 	&0.86	\\
SVR	&0.14 	&0.10 	&0.47 	&0.34 	&0.83 	\\
PCA	&0.14 	&0.09 	&0.47 	&0.63 	&0.82 	\\
RF	&0.17 	&0.12 	&0.52 	&0.48 	&0.77 	\\
MI	&0.18 	&0.12 	&0.56 	&0.55 	&0.73 	\\
ANOVA	&0.18 	&0.13 	&0.61 	&0.65 	&0.70 		\\
    \hline
\end{tabular}
\newline
  \caption{\color{black}Comparison of metrics for the observables selected from the 5 proposed feature selection algorithms using the MIDS suitability metric. The 4 wrapper methods described in Subsec.~\ref{subsec:FS} are included for comparison. The order of the methods are determined using the suitability metric (J). The other metrics (CC,rMSE, MAE, and rMAE) are as defined in Subsec.~\ref{subsec:FS}.}
 \label{tab:results_MOS}
\end{table}

\subsection{The best features for Disease Progression and Reconstruction (MIDS)}\label{subsec:MIDS}


{\color{black}Finally, we consider selecting obsevables  which optimize suitability with respect to  Minimal Immune and Disease State (MIDS) as defined in Subsec.~\ref{subsec:suitability_metrics}. These observables are optimal for predicting a combination of the Disease Progression Score (DPS) and all markers not included in the chosen set of 10 observables.

{\bf Overall Performance of FS Algorithms}:
In Table~\ref{tab:results_MOS} we rank  the proposed feature selection algorithms by performance with respect to the MIDS suitability metric, $J$ (as defined in Optimization Problem~\eqref{opt:FSProblem}).  We also report the other metrics (as defined in Eqs.~\eqref{eqn:CC}-\eqref{eqn:MAE}) and filter-based feature selection algorithms.

As in the MIS and MDS case, the results indicate that Sequential Forward Selection (SFS) based algorithms performed significantly better than embedded and filter methods with respect to all metrics.
 SFS wrapper method with Decision Tree Algorithms demonstrate the best performance according to the MIDS suitability metric.  All algorithms selected the maximum number of 10 allowable features. }

{\bf Most Important Features Using the SFS Algorithms}:
In Fig.~\ref{fig:dotplotMOS} we show the features that were selected by each of the proposed algorithms.  If we consider markers specific to helper and regulatory cells, the following features were each chosen by at least three of the five algorithms.

\begin{enumerate}
\item[(1)] \textit{CD4+GATA3+CD44+CD62L(Lo)} (4 times)
\item[(2)] \textit{CD4+Tbet+Ki67+} (4 times)
\item[(3)] \textit{CD4+Tbet+{\color{black}bc2+}} (4 times)
\item[(4)] \textit{CD4+GATA3+Ki67+{\color{black}bc2+}} (3 times)
\end{enumerate}

%

\section{Interpretation of Results}\label{sec:discussion}
{\color{black}
Here we summarize the results for each proposed metric of suitability: MDS, MIS, and MIDS.

{\bf Features for Predicting Disease Progression (MDS)}:
The MDS case is motivated by the need for T cell markers which have high accuracy when predicting disease progression and treatment outcome. In this context, we make the following observations.

The location feature was not selected by the top performing feature selection algorithms -- suggesting that the location where the T cells were collected is inconsequential to predicting the disease state. This implies that there is significant uniformity in the disease state among the lymph nodes and spleen.

In addition, 3 of 5 algorithms selected one antigen specific observable (\textit{CD4+FoxP3+CD25+Ki67+bc2+)} - indicating that the other selected T cell markers are likely correlated to autoimmune disease in general and are not sepecific to CIA.

Finally, we note that most of the selected biomarkers only consisted of 2 or 3 protein labels (unlike the more specific sub-populations selected in the MIS and MIDS cases).  This suggests that the ability to predict of disease progression and immunotherapy outcome is more robust (less prone to error) -- being based mostly on a well-established set of observables with larger sample sizes.

{\bf Features for Reconstructing Discarded Features (MIS)}:
The MIS case is motivated by the desire to reduce the number of markers used in flow cytometry by eliminating markers whose values can be inferred using a lower dimensional set of observables. However, the observables selected in this case are not necessarily correlated with disease progression or the effect of immunotherapy.

Because the MIS case selects markers which are not necessarily correlated with disease progression, a larger set of observables was selected and these observables generally include more peptide labels than in the MDS case.

Interestingly, unlike for MDS, relatively few regulatory T cell markers were selected in the MIS and MIDS case. This is likely because for these cases we include treatment as a potential observable. For MIS and MIDS, all algorithms now select treatment and this likely acts as a more reliable proxy for the regulatory population. This suggests that some caution is advised when deciding whether to include treatment in the set of selectable features.


Nearly all algorithms selected the location the T cells were collected as an important observable for predicting T cell populations.  This implies that many aspects of the immune state are not uniform across the lymph nodes and spleen.
}
{\color{black}
{\bf Features for Disease Progression and Reconstruction (MIDS)}:
The MIDS case combines the suitability metrics for MDS and MIS. Because weighting of the DPS score was relatively low, this case selected many of the same features as MIS.


Finally, we note that the memory T cell sub-population \textit{CD4+GATA3+CD44+CD62L(Lo)} was selected in all three cases (MDS, MIS, and MIDS).  It is clear that this sub-population is significant to both the immune and disease states.
}

\section{Conclusion}\label{sec:conclusions}
{\color{black}In this paper, we have considered the problem of using machine learning and feature selection algorithms to identify low dimensional subsets of observables (T cell markers) which are most useful in predicting disease progression and overall immune state.  Specifically, we have used a robust dataset of T cell markers obtained from mouse-model  immunotherapy for collagen induced arthritis. Moreover, we have identified the markers (Tables~\ref{tab:results_MDS}-\ref{tab:results_MOS}) which are most associated with the process of self-nonself determination. The algorithms proposed in this paper are general in that they can be used to identify lower dimensional subsets from any similar dataset. Furthermore, these algorithms have been made open source and are available for download online. Finally, we note that the list of biomarkers used in this set of experiments is not exhaustive and the accuracy of the results may be improved by testing whether inclusion of additional markers or exclusions (e.g. CD127+/-) alters the set of observables selected.}


%

\appendices
\section*{Funding}
This material is based upon work supported by the National Science Foundation under grant No. 1931270 and No. 1935453, and the National Institutes of Health under grants R01GM144966, R01AR078343, and R01AI155907.

\section*{Ethics approval and consent to participate}
The Institutional Animal Care and Use Committee (IACUC) of Arizona State University approved animal studies regarding the rheumatoid arthritis protocol number: 22-1904R. Approval date is 02/24/22. All animal experiments were conducted in accordance with the guidelines of Arizona State University and IACUC.

%
%
%




%
\bibliographystyle{IEEEtran}

\bibliography{IEEEJBHI}

\begin{thebibliography}{10}
\providecommand{\url}[1]{#1}
\csname url@samestyle\endcsname
\providecommand{\newblock}{\relax}
\providecommand{\bibinfo}[2]{#2}
\providecommand{\BIBentrySTDinterwordspacing}{\spaceskip=0pt\relax}
\providecommand{\BIBentryALTinterwordstretchfactor}{4}
\providecommand{\BIBentryALTinterwordspacing}{\spaceskip=\fontdimen2\font plus
\BIBentryALTinterwordstretchfactor\fontdimen3\font minus
  \fontdimen4\font\relax}
\providecommand{\BIBforeignlanguage}[2]{{%
\expandafter\ifx\csname l@#1\endcsname\relax
\typeout{** WARNING: IEEEtran.bst: No hyphenation pattern has been}%
\typeout{** loaded for the language `#1'. Using the pattern for}%
\typeout{** the default language instead.}%
\else
\language=\csname l@#1\endcsname
\fi
#2}}
\providecommand{\BIBdecl}{\relax}
\BIBdecl

\bibitem{fisher2020situ}
J.~Fisher, W.~Zhang, S.~Balmert, A.~Aral, A.~Acharya, Y.~Kulahci, J.~Li,
  H.~Turnquist, A.~Thomson, M.~Solari \emph{et~al.}, ``In situ recruitment of
  regulatory t cells promotes donor-specific tolerance in vascularized
  composite allotransplantation,'' \emph{Science Advances}, vol.~6, no.~11,
  2020.

\bibitem{ratay2017treg}
M.~Ratay, A.~Glowacki, S.~Balmert, A.~Acharya, J.~Polat, L.~Andrews,
  M.~Fedorchak, J.~Schuman, D.~Vignali, and S.~Little, ``Treg-recruiting
  microspheres prevent inflammation in a murine model of dry eye disease,''
  \emph{Journal of Controlled Release}, 2017.

\bibitem{ratay2017tri}
M.~Ratay, S.~Balmert, A.~Acharya, A.~Greene, T.~Meyyappan, and S.~Little, ``Tri
  microspheres prevent key signs of dry eye disease in a murine, inflammatory
  model,'' \emph{Scientific reports}, vol.~7, no.~1, 2017.

\bibitem{acharya2017localized}
A.~Acharya, M.~Sinha, M.~Ratay, X.~Ding, S.~Balmert, C.~Workman, Y.~Wang,
  D.~Vignali, and S.~Little, ``Localized multi-component delivery platform
  generates local and systemic anti-tumor immunity,'' \emph{Advanced Functional
  Materials}, vol.~27, no.~5, 2017.

\bibitem{jaggarapu2023orally}
M.~M. C.~S. Jaggarapu, A.~Thumsi, R.~Nile, B.~D. Ridenour, T.~Khodaei, A.~P.
  Suresh, A.~Esrafili, K.~Jin, and A.~P. Acharya, ``Orally delivered {2D}
  covalent organic frameworks releasing kynurenine generate anti-inflammatory
  {T} cell responses in collagen induced arthritis mouse model,''
  \emph{Biomaterials}, p. 122204, 2023.

\bibitem{mangal2021immunometabolism}
J.~L. Mangal, N.~Basu, H.-J.~J. Wu, and A.~P. Acharya, ``{I}mmunometabolism:
  {A}n {E}merging {T}arget {F}or {I}mmunotherapies to {T}reat {R}heumatoid
  {A}rthritis,'' \emph{Immunometabolism}, vol.~3, no.~4, 2021.

\bibitem{brodin2017human}
P.~Brodin and M.~Davis, ``Human immune system variation,'' \emph{Nature reviews
  immunology}, vol.~17, no.~1, 2017.

\bibitem{davis2008prescription}
M.~Davis, ``A prescription for human immunology,'' \emph{Immunity}, vol.~29,
  no.~6, 2008.

\bibitem{germain2011human}
R.~Germain and P.~Schwartzberg, ``The human condition: an immunological
  perspective,'' \emph{Nature immunology}, vol.~12, no.~5, 2011.

\bibitem{mangal2020metabolite}
J.~Mangal, S.~Inamdar, Y.~Yang, S.~Dutta, M.~Wankhede, X.~Shi, H.~Gu, M.~Green,
  K.~Rege, M.~Curtis \emph{et~al.}, ``Metabolite releasing polymers control
  dendritic cell function by modulating their energy metabolism,''
  \emph{Journal of Materials Chemistry B}, vol.~8, no.~24, 2020.

\bibitem{jaggarapu2023alpha}
M.~M. Jaggarapu, D.~Ghosh, T.~Johnston, J.~R. Yaron, J.~L. Mangal, S.~Inamdar,
  M.~Gosangi, K.~Rege, and A.~P. Acharya, ``Alpha-ketoglutaric acid based
  polymeric particles for cutaneous wound healing,'' \emph{Journal of
  Biomedical Materials Research Part A}, 2023.

\bibitem{inamdar2023biomaterial}
S.~Inamdar, T.~Tylek, A.~Thumsi, A.~P. Suresh, M.~M. C.~S. Jaggarapu, M.~Halim,
  S.~Mantri, A.~Esrafili, N.~D. Ng, E.~Schmitzer \emph{et~al.}, ``Biomaterial
  mediated simultaneous delivery of spermine and alpha ketoglutarate modulate
  metabolism and innate immune cell phenotype in sepsis mouse models,''
  \emph{Biomaterials}, vol. 293, p. 121973, 2023.

\bibitem{mangal2022short}
J.~L. Mangal, S.~Inamdar, A.~P. Suresh, M.~M. C.~S. Jaggarapu, A.~Esrafili,
  N.~D. Ng, and A.~P. Acharya, ``Short term, low dose alpha-ketoglutarate based
  polymeric nanoparticles with methotrexate reverse rheumatoid arthritis
  symptoms in mice and modulate {T} helper cell responses,'' \emph{Biomaterials
  Science}, vol.~10, no.~23, pp. 6688--6697, 2022.

\bibitem{thumsi2023vaccines}
A.~Thumsi, S.~J. Swaminathan, J.~L. Mangal, A.~P. Suresh, and A.~P. Acharya,
  ``Vaccines {P}revent {R}einduction of {R}heumatoid {A}rthritis {S}ymptoms in
  {C}ollagen-{I}nduced {A}rthritis {M}ouse {M}odel,'' \emph{Drug Delivery and
  Translational Research}, vol.~13, no.~7, pp. 1925--1935, 2023.

\bibitem{mangal2021inhibition}
J.~Mangal, S.~Inamdar, T.~Le, X.~Shi, M.~Curtis, H.~Gu, and A.~Acharya,
  ``Inhibition of glycolysis in the presence of antigen generates suppressive
  antigen-specific responses and restrains rheumatoid arthritis in mice,''
  \emph{Biomaterials}, vol. 277, 2021.

\bibitem{jmlr}
B.~Colbert and M.~Peet, ``A convex parametrization of a new class of universal
  kernel functions,'' \emph{Journal of Machine Learning Research}, vol.~21,
  no.~45, 2020.

\bibitem{colbert2020new}
------, ``A new algorithm for tessellated kernel learning,'' 2020.

\bibitem{Chandrashekar2014ASO}
G.~Chandrashekar and F.~Sahin, ``A survey on feature selection methods,''
  \emph{Computers and Electrical Engineering}, vol.~40, no.~1, 2014.

\bibitem{dash1997feature}
M.~Dash and H.~Liu, ``Feature selection for classification,'' \emph{Intelligent
  data analysis}, vol.~1, 1997.

\bibitem{Battiti1994UsingMI}
R.~Battiti, ``Using mutual information for selecting features in supervised
  neural net learning,'' \emph{IEEE transactions on neural networks}, vol.~5,
  no.~4, 1994.

\bibitem{kumar2015feature}
M.~Kumar, N.~K. Rath, A.~Swain, and S.~Rath, ``Feature selection and
  classification of microarray data using mapreduce based anova and k-nearest
  neighbor,'' \emph{Procedia Computer Science}, vol.~54, 2015.

\bibitem{Song2010FeatureSU}
F.~Song, Z.~Guo, and D.~Mei, ``Feature selection using principal component
  analysis,'' \emph{2010 International Conference on System Science,
  Engineering Design and Manufacturing Informatization}, vol.~1, 2010.

\bibitem{Guyon2004GeneSF}
I.~Guyon, J.~Weston, S.~Barnhill, and V.~Vapnik, ``Gene selection for cancer
  classification using support vector machines,'' \emph{Machine Learning},
  vol.~46, 2004.

\bibitem{Zhang2006RecursiveSF}
X.~Zhang, X.~Lu, Q.~Shi, X.~q~Xu, H.~Leung, L.~Harris, J.~Iglehart, A.~Miron,
  J.~Liu, and W.~Wong, ``Recursive svm feature selection and sample
  classification for mass-spectrometry and microarray data,'' \emph{BMC
  Bioinformatics}, vol.~7, 2006.

\bibitem{Jong2004FeatureSI}
K.~Jong, E.~Marchiori, M.~Sebag, and A.~W. Vaart, ``Feature selection in
  proteomic pattern data with support vector machines,'' \emph{2004 Symposium
  on Computational Intelligence in Bioinformatics and Computational Biology},
  2004.

\bibitem{Prados2004MiningMS}
J.~Prados, A.~Kalousis, J.~Sanchez, L.~Allard, O.~Carrette, and M.~Hilario,
  ``Mining mass spectra for diagnosis and biomarker discovery of cerebral
  accidents,'' \emph{Proteomics}, vol.~4, no.~8, 2004.

\bibitem{Breiman1984ClassificationAR}
L.~Breiman, J.~Friedman, R.~Olshen, and C.~J. Stone, ``Classification and
  regression trees,'' \emph{Biometrics}, vol.~40, no.~3, 1984.

\bibitem{Borchani2015ASO}
H.~Borchani, G.~Varando, C.~Bielza, and P.~Larra{\~n}aga, ``A survey on
  multi‐output regression,'' \emph{Wiley Interdisciplinary Reviews: Data
  Mining and Knowledge Discovery}, vol.~5, no.~5, 2015.

\bibitem{colbert2020convex}
B.~Colbert and M.~Peet, ``A convex parametrization of a new class of universal
  kernel functions,'' \emph{Journal of Machine Learning Research}, vol.~21,
  2020.

\bibitem{githubMS}
A.~Talitckii, B.~K. Colbert, and M.~M. Peet, ``Minimal {I}mmune and {D}isease
  {S}tates,'' \url{https://github.com/CyberneticSCL/JBHI23_MinimalStates},
  2022.

\end{thebibliography}
\vspace{-2mm}
\vspace{-2mm}
\vspace{-2mm}
\vspace{-2mm}
\begin{IEEEbiography}[{\includegraphics[width=1in,height=1.25in,clip,keepaspectratio]{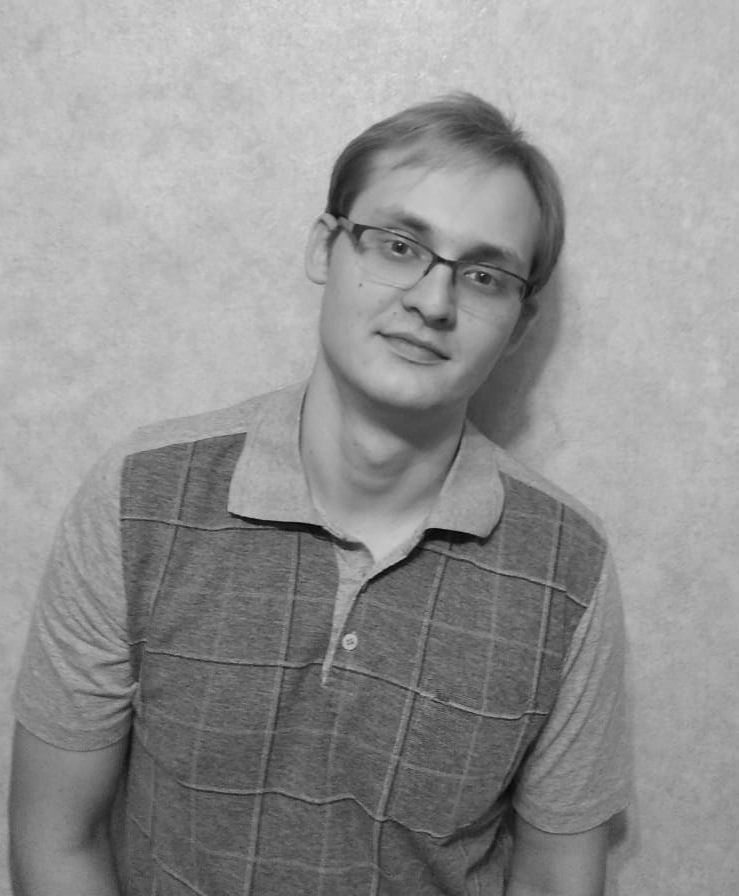}}]{Aleksandr Talitckii}
 received the bachelor’s degree in high energy physics from the Moscow Institute of Physics and Technology, Dolgoprudny, Russia, and the M.Sc. degree in information science and technology from the Skolkovo Institute of Science and Technology (Skoltech), Moscow, Russia, in 2020. In 2019, he was an Intern with the Fondazione Bruno Kessler Research Center, Povo, Italy. He is currently pursuing the Ph.D. degree with Arizona State University, Tempe, AZ, USA.  His research interests include machine learning, control theory and medical related applications.
\end{IEEEbiography}

\vspace{-2mm}
\vspace{-2mm}
\vspace{-2mm}
\begin{IEEEbiography}[{\includegraphics[width=1in,height=1.25in,clip,keepaspectratio]{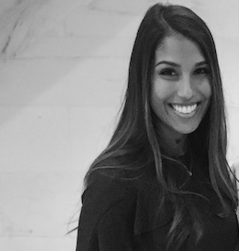}}]{Joslyn L. Mangal}
 received her PhD at Arizona State University in Biological
Design in 2022. Her doctoral research resided in the fields of immunoengineering,
biomaterials and drug delivery. Specifically, her work focused on constructing
biomaterials to alter immune cell function and behavior to provide solutions for
autoimmune rheumatoid arthritis. Joslyn’s long-term research goals are to contribute to
work involved in furthering our understanding of how gut health influences the
development and progression of immune-mediated diseases.
\end{IEEEbiography}

\vspace{-2mm}
\vspace{-2mm}
\vspace{-2mm}
\begin{IEEEbiography}[{\includegraphics[width=1in,height=1.25in,clip,keepaspectratio]{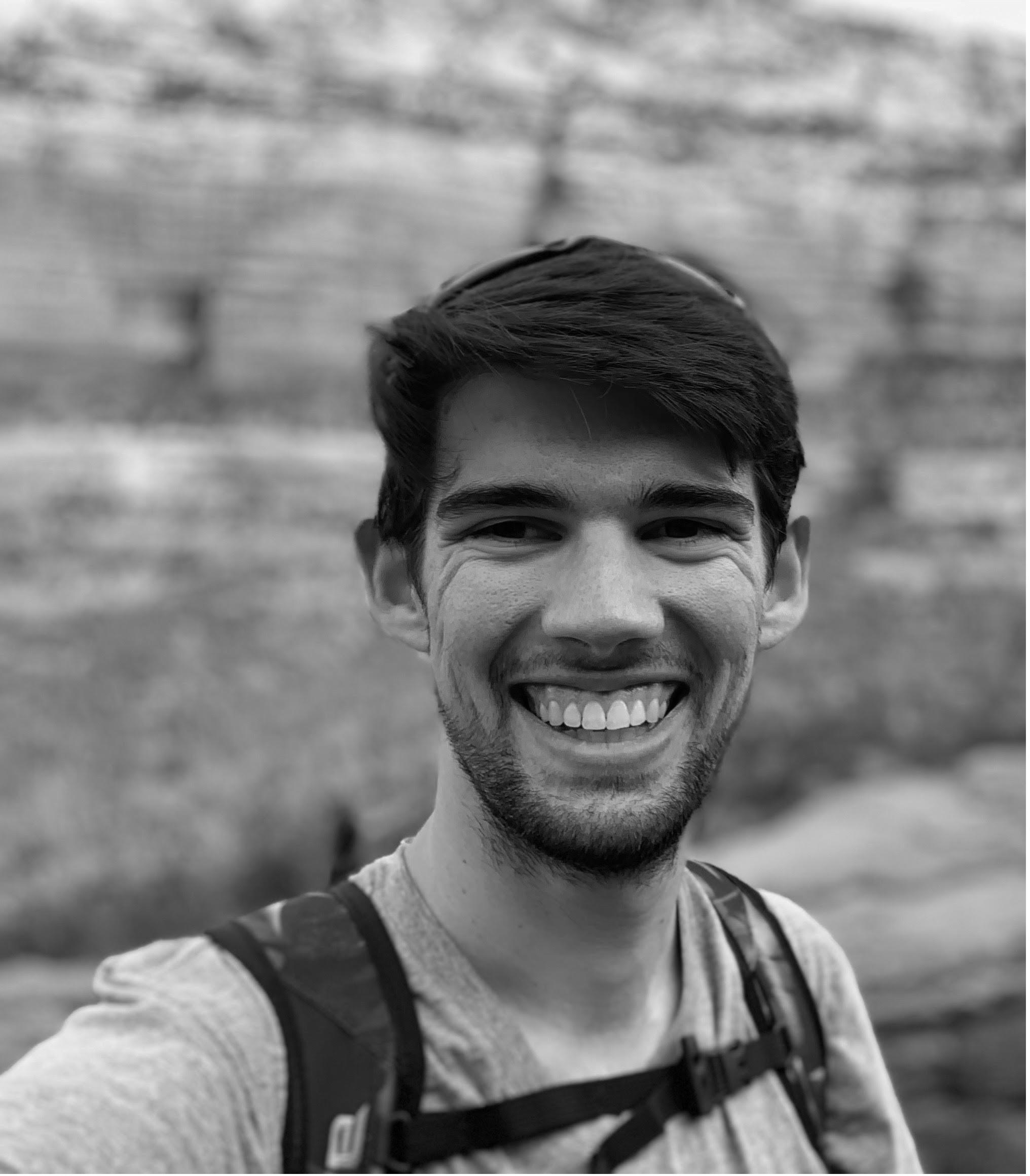}}]{Brendon K. Colbert}
 received his PhD degree in mechanical engineering at Arizona State University in 2021 studying control systems engineering, the immune system and machine learning.  After graduation he began work at NASA as an Aerospace Research Engineer.  He currently works at Apple as a product design engineer.
\end{IEEEbiography}

\vspace{-2mm}
\vspace{-2mm}
\vspace{-2mm}
\vspace{-2mm}
\begin{IEEEbiography}[{\includegraphics[width=1in,height=1.25in,clip,keepaspectratio]{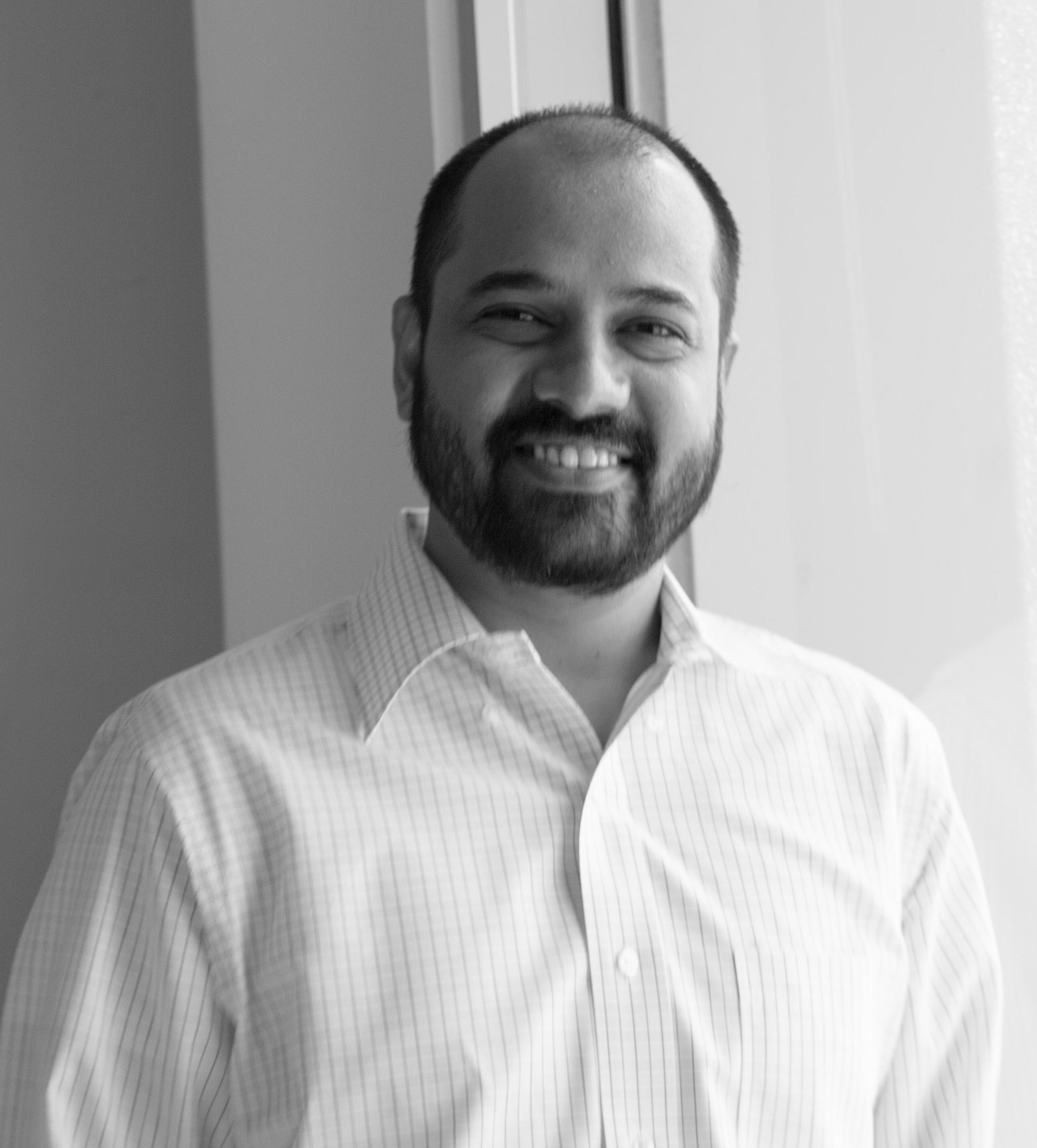}}]{Abhinav P. Acharya}
 received his PhD degree in Materials Science and Engineering at the University of Florida in 2010 focused on developing vaccines for type 1 diabetes. He started his independent research position in the Department of Chemical Engineering at Arizona State University in 2019, after doing postdoctoral fellowships at the University of California, Berkeley, and the University of Pittsburgh. In this current role at Arizona State University, he leads teams spanning a range of different studies on immunoengineering, biomaterials and drug delivery.
\end{IEEEbiography}

\vspace{-2mm}
\vspace{-2mm}
\vspace{-2mm}
\begin{IEEEbiography}[{\includegraphics[width=1in,height=1.25in,clip,keepaspectratio]{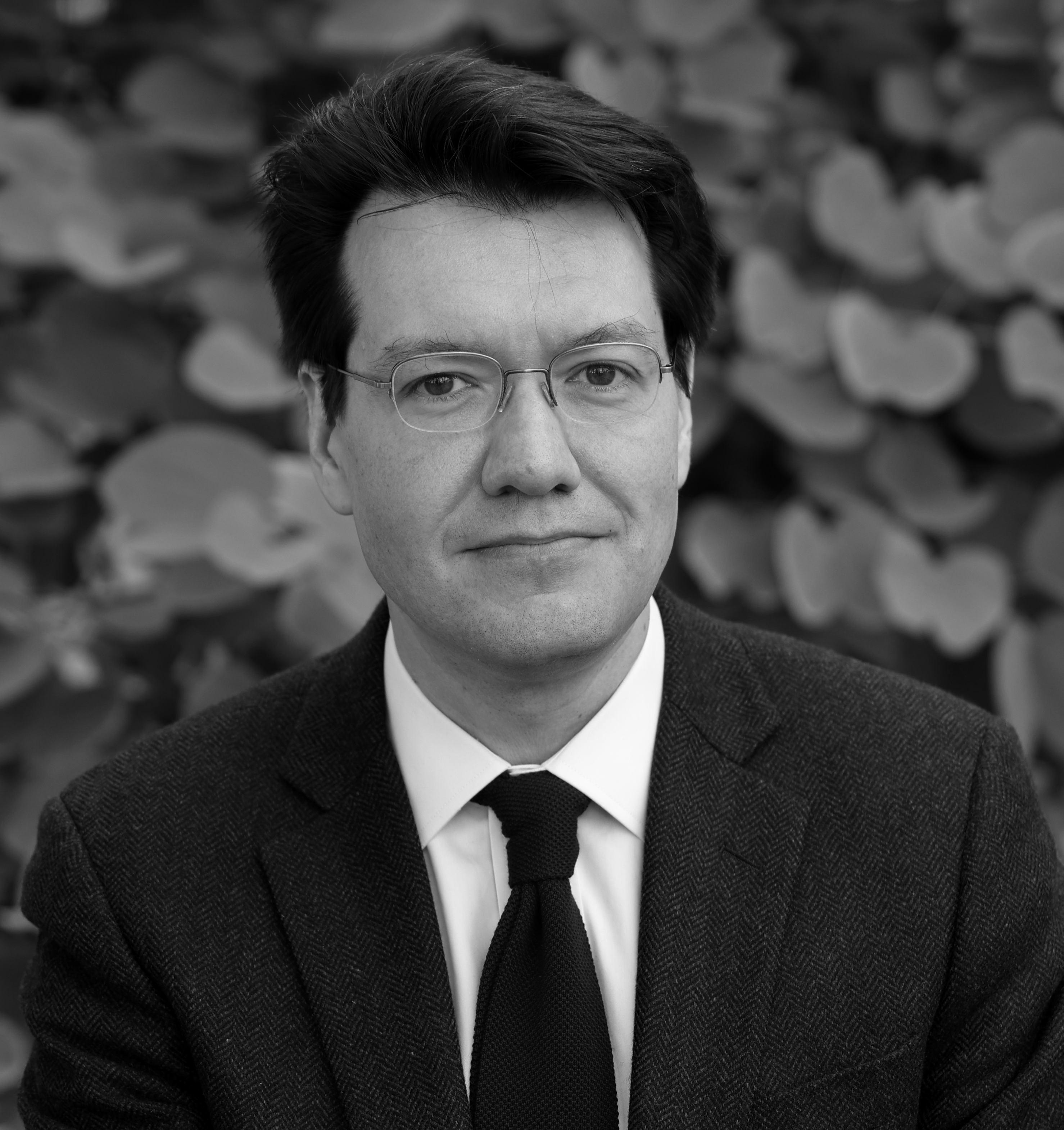}}]{Matthew M. Peet}
 received the B.S. degree in physics and in aerospace engineering from the University of Texas, Austin, TX, USA, in 1999, and the M.S. and Ph.D. degrees in aeronautics and astronautics from Stanford University, Stanford, CA, USA, in 2001 and 2006, respectively. He was a Postdoctoral Fellow with INRIA, Paris, France, from 2006 to 2008. He was an Assistant Professor of aerospace engineering with the Illinois Institute of Technology, Chicago, IL, USA, from 2008 to 2012. He is currently an Associate Professor of aerospace engineering with Arizona State University, Tempe, AZ, USA.,Dr. Peet was the recipient of a National Science Foundation CAREER Award in 2011.
\end{IEEEbiography}

\end{document}